\pdfoutput=1

\documentclass[11pt]{article}

\usepackage[final]{acl}

\usepackage{times}
\usepackage{latexsym}

\usepackage[T1]{fontenc}

\usepackage[utf8]{inputenc}

\usepackage{microtype}

\usepackage{inconsolata}

\usepackage{graphicx}
\usepackage{algorithm}
\usepackage{tabularx}
\usepackage{colortbl}
\usepackage{amssymb}
\usepackage{subfigure}
\usepackage{multirow}
\usepackage{algpseudocode}  
\usepackage{makecell}
\usepackage{arydshln}

\title{Enhancing Rumor Detection Methods with Propagation Structure Infused Language Model}

\author{
 \textbf{Chaoqun Cui},
 \textbf{Siyuan Li},
 \textbf{Kunkun Ma},
 \textbf{Caiyan Jia\thanks{Corresponding author.}}
\\
\\
School of Computer Science and Technology \& Beijing Key Lab of Traffic Data\\ Analysis and Mining Beijing Jiaotong University, Beijing 100044, China
\\
 \texttt{\{ccqun19990728,pratearon\}@gmail.com}\\
 \texttt{\{siyuanli,cyjia\}@bjtu.edu.cn}
}

\begin{document}
\maketitle
\begin{abstract}

Pretrained Language Models (PLMs) have excelled in various Natural Language Processing tasks, benefiting from large-scale pretraining and self-attention mechanism's ability to capture long-range dependencies. However, their performance on social media application tasks like rumor detection remains suboptimal. We attribute this to mismatches between pretraining corpora and social texts, inadequate handling of unique social symbols, and pretraining tasks ill-suited for modeling user engagements implicit in propagation structures. To address these issues, we propose a continue pretraining strategy called Post Engagement Prediction (PEP) to infuse information from propagation structures into PLMs. PEP makes models to predict root, branch, and parent relations between posts, capturing interactions of stance and sentiment crucial for rumor detection. We also curate and release large-scale Twitter corpus: TwitterCorpus (269GB text), and two unlabeled claim conversation datasets with propagation structures (UTwitter and UWeibo). Utilizing these resources and PEP strategy, we train a Twitter-tailored PLM called SoLM. Extensive experiments demonstrate PEP significantly boosts rumor detection performance across universal and social media PLMs, even in few-shot scenarios. On benchmark datasets, PEP enhances baseline models by 1.0-3.7\% accuracy, even enabling it to outperform current state-of-the-art methods on multiple datasets. SoLM alone, without high-level modules, also achieves competitive results, highlighting the strategy's effectiveness in learning discriminative post interaction features.

\end{abstract}

\section{Introduction}

Recent years have seen Pretrained Language Models (PLMs) based on Transformer \cite{bert,gpt2,gpt3} excel in various Natural Language Processing (NLP) tasks like machine translation \cite{transformer,trans,sspo}, sentiment analysis \cite{sa1,sa2}, and question-answering systems \cite{bert,xlnet}. The success is largely due to the parallel computation of self-attention mechanism, enabling long-range dependency capture in texts and intricate semantic learning. Furthermore, PLMs benefit from large-scale pretraining on unlabeled corpora with increased model capacity and depth. In specialized domains, pretraining with extensive unlabeled professional corpora \cite{med,law,fin} allows models to absorb domain-specific knowledge and concepts, thus improving performance on related tasks.

The text in social media platforms like Weibo and Twitter originates from user-generated posts and comments. This type of corpus, differing from most text corpora used for language model pretraining (such as book corpora, Wikipedia corpora, etc.), tends to be shorter, highly emotive, and possesses explicit directional relations. In other words, texts on social media contain interactions among users, where comments are specifically directed at other users' posts or comments. Current universal PLMs predominantly utilize token-level pretraining tasks like Causal Language Modeling (CLM) \cite{gpt}, which evidently struggle to model text interaction. 
In this study, we focus on rumor detection, a typical social media application task, to explore how to enhance the universal PLMs performance in such applications.

In prevalent rumor detection methods \cite{bigcn,ragcl}, learning from propagation structures of claims (a claim refers to the source post and its comments) is a common strategy, emphasizing semantics, stance, sentiment, and post/user interactions. However, these strategies have not significantly benefited from prevalent universal PLMs. This is reflected in the limited performance boost in rumor detection models employing universal PLMs for initial feature extraction, as compared to traditional methods like word2vec \cite{word2vec} and tf-idf \cite{tfidf}. Universal PLMs and word2vec lack inherent understanding of sentiment and stance engagements among posts, necessitating further training via high-level models such as GNNs. 

We investigate this underperformance of universal PLMs. To enhance their performance in rumor detection, we propose a continue pretraining strategy called Post Engagement Prediction (PEP). PEP aims to integrate user engagement information, inherent in propagation structures, into PLMs. Additionally, we have collected and open-sourced high-quality data resources, including a large-scale Twitter corpus named TwitterCorpus and two large-scale conversation dataset with propagation structures called Unlabeled Twitter (UTwitter) and Unlabeled Weibo (UWeibo). Using these corpora and PEP strategy, we trained a BERT architecture PLM tailored for social media application tasks (for Twitter platform), named Social Language Model (SoLM). We believe PEP can not only improve PLMs' performance in rumor detection but also offer insights for other social media application tasks such as content recommendation, social network analysis, and user behavior analysis.

In summary, this study contributes as follows:
\begin{itemize}
\item We ran extensive experiments, demonstrating the poor performance of universal PLMs in rumor detection and analyzing the reasons.
\item We proposed the PEP strategy to integrate user interaction information into PLMs.
\item We collected multiple corpora and trained SoLM. We released all our resources.
\item Experiments indicate that PLMs trained with PEP enhance the performance of existing rumor detection methods, with even more pronounced improvements in few-shot scenarios.
\end{itemize}

\section{Related Work}

In this section, we will review the related works.

\subsection{Rumor Detection}

Among the existing studies, early rumor detection methods mainly take advantage of traditional classification methods by using hand-crafted features \cite{dtc,rfc,feature1}. Deep learning has greatly promoted the development of rumor detection methods. These approaches generally fall into four categories: time-series based techniques \cite{yucnn,user1,liuandwu} modeling text content or user profiles as time series; propagation structure learning methods \cite{rvnn,sog,ebgcn,debiased,adgscl,urumor} accounting for propagation structures of initial rumors and their replies; multi-source integration approaches \cite{ms1,ms3,ms2} combining various rumor resources, such as post content, user profiles, and relations between posts and users; and multi-modal fusion techniques \cite{otherrumor1,eann,mm1,vga} that use both post content and associated images for efficient rumor debunking.

In the literature, the significance of propagation structure has been increasingly recognized. Numerous state-of-the-art (SOTA) models employ GNNs to model propagation trees. BiGCN \cite{bigcn} implemented a bidirectional Graph Convolutional Network (GCN) \cite{gcn} along with a root node feature enhancement technique. PLAN \cite{plan} established a Transformer cognizant of the propagation tree structures. ClaHi-GAT \cite{clahi} used GAT on undirected graphs with sibling relations to model user interactions. GACL \cite{gacl} employed contrastive loss with adversarial training to learn noise-resilient representations of rumors. RAGCL \cite{ragcl} designed an adaptive graph contrastive learning method considering the structural characteristics of propagation trees. Together, these studies highlight the crucial role of propagation structures and post texts. 

\subsection{Social Media Language Models}

There exists various language models specifically designed for social media. For instance, BERTweet \cite{bertweet} replicated RoBERTa on 850 million tweets. TimeLMs \cite{timelms} utilized a set of RoBERTa models \cite{roberta} to learn from English tweets across various time ranges. Another example is XLM-T \cite{xlmt}, which extended the pretraining process from a XLM-R checkpoint \cite{xlmr} utilizing 198 million multilingual tweets. Additionally, TwHIN-BERT \cite{twhin} models user engagements as a heterogeneous graph and then utilizes user interaction information on the graph during training process. These PLMs model texts from social corpora, alleviating some issues of universal PLMs in rumor detection. However, their pretraining methods overlook the learning of post engagements and semantic association between multiple posts characterized by propagation structures, which are crucial for rumor identification. 

\section{Problem Analysis}

In this section, we discuss the suboptimal performance exhibited by universal PLMs and its reasons.

\subsection{Inefficacy of Universal PLMs}

The claim propagation process follows a tree structure, with source post as root and comments as other nodes. The reply relation between comments serve as edges. This tree is the primary data structure processed by rumor detection methods based on propagation structure.
See Appendix~\ref{sec:rpt} for examples of propagation tree. Typically, the interaction relation between comments of rumor and non-rumor claims is markedly different. This is manifested specifically as comments to rumor claims having more intense stances and sentiments, while those to non-rumor claims tend to be more moderate. Propagation structure based methods focus on learning stance and sentiment interaction among posts. These methods generally use common text feature extraction methods for initial post feature vectors, which are then processed by high-level models like GNNs to learn inter-post relations. 

\begin{table*}[h]
\centering
\resizebox{1.0\textwidth}{!}{
\begin{tabular}{cccccccc}
\Xhline{1.0pt}
\rowcolor{gray!20}
\textbf{Statistic} & \textbf{Weibo} & \textbf{DRWeibo} & \textbf{Twitter15} & \textbf{Twitter16} & \textbf{PHEME} & \textbf{UWeibo}  & \textbf{UTwitter}\\
\hline
\textbf{language} & zh & zh & en & en & en & zh & en \\
\textbf{labeled} & True & True & True & True & True & False & False \\
\textbf{\# claims} & 4664 & 6037 & 1490 & 818 & 6425 & 209549 & 204922 \\
\textbf{\# non-rumors} & 2351 & 3185 & 374 & 205 & 4023 & - & - \\
\textbf{\# false rumors} & 2313 & 2852 & 370 & 205 & 638 & - & - \\
\textbf{\# true rumors} & - & - & 372 & 207 & 1067 & - & -\\
\textbf{\# unverified rumors} & - & - & 374 & 201 & 697 & - & - \\
\textbf{avg num posts} & 803.5 & 61.8 & 31.1 & 25.9 & 15.4 & 50.5 & 82.5\\
\Xhline{1.0pt}
\end{tabular}
}
\caption{Statistics of the datasets.}
\label{tab:sta}
\end{table*}

\begin{table*}[h]
\centering
\resizebox{1.0\textwidth}{!}{
\begin{tabular}{cccccccc}
 \Xhline{1.0pt}
 \rowcolor{gray!20}
 ~ & ~ & ~ & \multicolumn{5}{c}{\textbf{Dataset}}\\
 \cline{4-8}
 \rowcolor{gray!20}
 \multirow{-2}{*}{\textbf{Method}} & \multirow{-2}{*}{\textbf{Initialization}} & \multirow{-2}{*}{\textbf{Parameters}} & \textbf{Weibo} & \textbf{DRWeibo} & \textbf{Twitter15} & \textbf{Twitter16} & \textbf{PHEME}\\
 \hline
  \multirow{8}{*}{\textbf{PLAN}} & TF-IDF & - & 90.8 & 74.3 & 80.2 & 82.0 & 65.3 \\
 ~ & Word2Vec & - & 91.5 & 78.8 & 81.9 & 84.3 & 68.6 \\
 \cdashline{2-8}
 ~ & BERT & 110M & 91.2 & 77.9 & 82.7 & 83.7 & 68.7 \\
 ~ & RoBERTa & 125M & 91.8 & 78.3 & 82.4 & 83.0 & 67.8 \\
 ~ & BERTweet & 110M & - & - & 83.2 & \textbf{84.5} & 68.5 \\
 ~ & TwHIN-BERT & 280M & - & - & 82.8 & 84.3 & 69.5 \\
 \cdashline{2-8}
 ~ & Baichuan2 & 7B & \textbf{92.5} & \textbf{79.4} & - & - & - \\
 ~ & LLaMA2 & 7B & - & - & \textbf{83.4} & 84.0 & \textbf{70.2} \\
 \hline
 \multirow{8}{*}{\textbf{BiGCN}} & TF-IDF & - & 93.1 & 84.2 & 81.8 & 84.7 & 66.7 \\
 ~ & Word2Vec & - & 94.2 & 86.6 & 84.4 & \textbf{88.0} & 70.8 \\
 \cdashline{2-8}
 ~ & BERT & 110M & \textbf{94.4} & 86.1 & 83.5 & 87.9 & 70.3 \\
 ~ & RoBERTa & 125M & 93.8 & 87.2 & 83.8 & 87.3 & 70.5 \\
 ~ & BERTweet & 110M & - & - & 84.9 & 87.8 & 71.2 \\
 ~ & TwHIN-BERT & 280M & - & - & \textbf{85.2} & 87.2 & 71.8 \\
 \cdashline{2-8}
 ~ & Baichuan2 & 7B & 94.0 & \textbf{88.7} & - & - & - \\
 ~ & LLaMA2 & 7B & - & - & 85.0 & 87.0 & \textbf{72.0} \\
 \hline
 \multirow{8}{*}{\textbf{GACL}} & TF-IDF & - & 92.8 & 85.7 & 84.9 & 85.9 & 66.9\\
 ~ & Word2Vec & - & 93.0 & \textbf{87.4} & 85.0 & 89.5 & 71.2\\
 \cdashline{2-8}
 ~ & BERT & 110M & 93.8 & 87.0 & 84.6 & 89.1 & 71.1\\
 ~ & RoBERTa & 125M & 93.4 & 86.4 & 85.3 & 89.4 & 70.3\\
 ~ & BERTweet & 110M & - & - & 85.5 & \textbf{90.2} & 71.7 \\
 ~ & TwHIN-BERT & 280M & - & - & 85.8 & 88.8 & 71.4 \\
 \cdashline{2-8}
 ~ & Baichuan2 & 7B & \textbf{94.3} & 87.1 & - & - & - \\
 ~ & LLaMA2 & 7B & - & - & \textbf{86.0} & 89.8 & \textbf{72.3} \\
 \Xhline{1.0pt}
\end{tabular}
}
\caption{The impact of feature initialization methods. BERT and RoBERTa are employed on Chinese and English datasets respectively, using their corresponding Chinese and English models.}
\label{tab:init}
\end{table*}

We examined the impact of feature initialization methods on rumor detection model including PLAN, BiGCN and GACL across five datasets: Weibo \cite{weibo}, DRWeibo \cite{ragcl}, Twitter15, Twitter16 \cite{twitter1516}, and PHEME \cite{pheme}. These datasets originate from two large platforms, Twitter and Weibo. We reported macro F1 score on the class-imbalanced dataset PHEME, and accuracy on the other class-balanced datasets. The dataset statistics and the results are presented in Table~\ref{tab:sta} and~\ref{tab:init}. In experiments, we involved traditional methods such as tf-idf and word2vec (skip-gram), as well as autoencoding language models like BERT, RoBERTa, BERTweet and TwHIN-BERT, and the generative large language model Baichuan2 \cite{baichuan2} and LLaMA2 \cite{llama2}. For Baichuan2 and LLaMA2, we utilized the embedding of the last token in a tweet as its representation.

The results indicate:
(1) Universal PLMs do not exhibit significant improvement over traditional methods, such as word2vec;
(2) Among autoencoding PLMs, TwHIN-BERT and BERTweet models pretrained on Twitter corpus outperforms universal PLMs in most scenarios;
(3) While SOTA generative large models (like Baichuan2 and LLaMA2) have several orders of magnitude more parameters compared to traditional PLMs, they do not lead to noticeably better performance. 
Given the outstanding performance of universal PLMs in other domains \cite{bert,gpt}, their suboptimal results in rumor detection becomes a question worth investigating. 

\subsection{Cause Analysis}

We attribute the underperformance of universal PLMs primarily to three factors. (1) The training corpora of universal PLMs do not align with social media texts. (2) Universal PLMs are not equipped to properly process symbols unique to social media texts. (3) The pretraining tasks employed by universal PLMs are ill-suited for rumor detection tasks. We will expound on these points.

\subsubsection{Training Corpus Mismatch}

Universal PLMs are usually trained on corpora such as books and articles (like BooksCorpus \cite{bookscorpus} or Project Gutenberg\footnote{\url{https://www.gutenberg.org}}), Wikipedia corpora, and web-crawled data (like Common Crawl\footnote{\url{https://commoncrawl.org}}), with language that is generally more formal, grammatically correct, and skewed towards the written form. However, texts in posts on social platforms is usually colloquial, expressive in a more spoken style, and includes uncivilized language, slang, abbreviations (like U, IC, OIC, THX), emojis, and unique internet terms. 

\begin{figure}[t]
  \centering
  \includegraphics[width=0.48\textwidth]{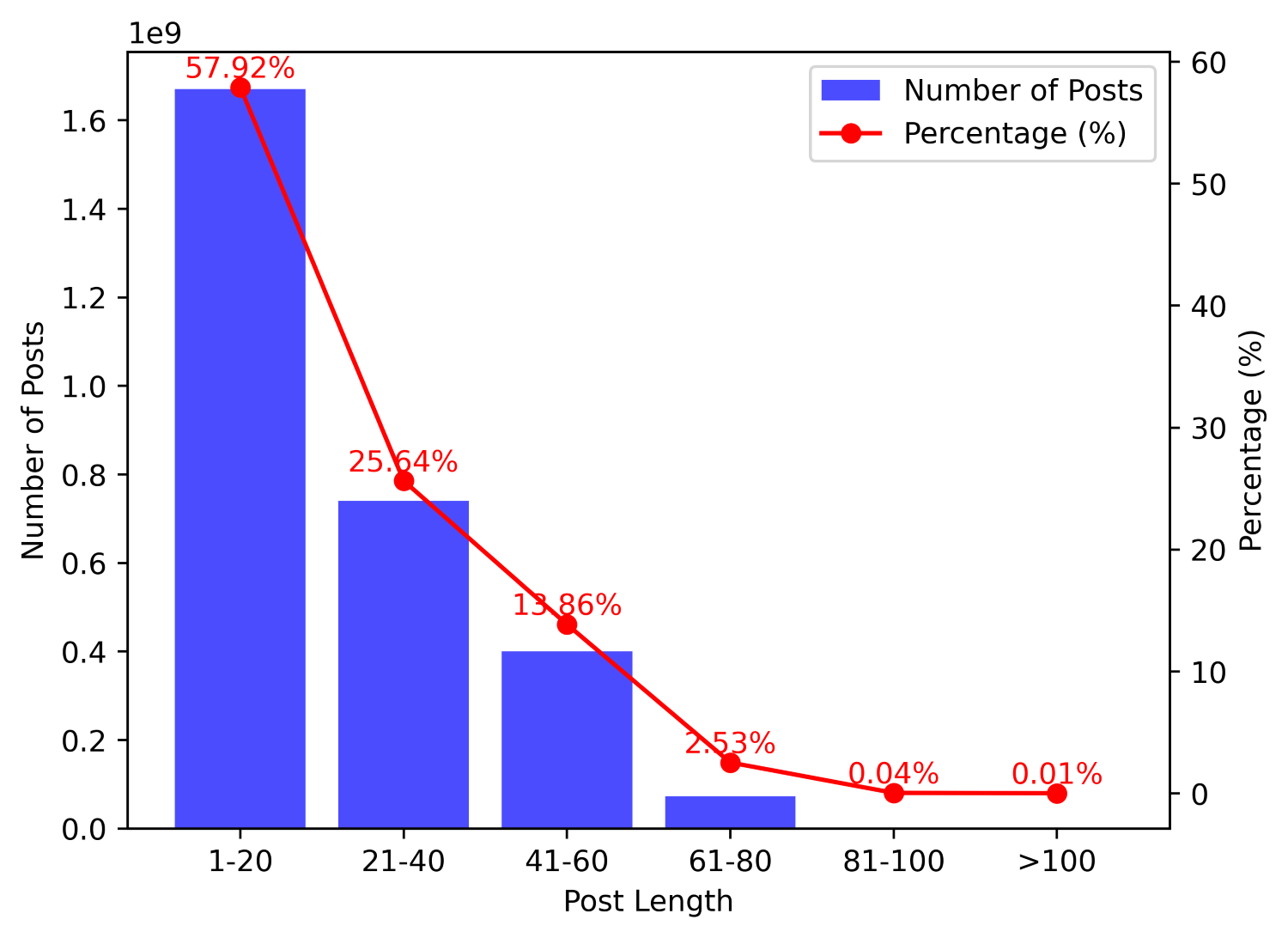}
  \caption{Post lengths distribution on TwitterCorpus.}
  \label{fig:pd}
\end{figure}

Universal PLMs are mainly trained on long texts, while social media posts are usually very short. We counted the length distribution of 2.8 billion tweets in TwitterCorpus, as shown in Figure~\ref{fig:pd}. We found that posts tend to be very brief, with most (57.92\%) having fewer than 20 tokens, and virtually none (0.01\%) exceeding 100 tokens. This indicates that the length distribution of texts from social media platforms is significantly different from corpora like BooksCorpus and Wikipedia. This may lead to problems for universal PLMs when dealing with short texts. First, although models pretrained on longer texts excel at capturing long-distance contextual relations, this strength may not be crucial for short texts, leading to potential misalignment with the characteristics of short-text tasks. Second, there could be disparities in vocabulary and grammatical features between long and short texts. For example, short texts (e.g., tweets, text messages) might contain more informal language, slang, emojis, and abbreviations (as mentioned before). If the PLMs haven't thoroughly learned these features, they could struggle with short texts.

\subsubsection{Symbol Processing Shortfalls} 
\label{sec:sps}

Social media posts contain special symbols that represent specific interactive behaviors, mainly including user mentions (like @someone), web/url links, topic tags (like \#Covid19), and emojis. These symbols may affect the text semantics learned by PLMs, while some universal PLMs (like BERT, RoBERTa) lack the ability to handle these special symbols properly. Some necessary processes include: (1) Mitigating the impact of user mentions and web links, which usually do not affect the text content; (2) Identifying topic tags in the texts, as they often delineate the subject matter of posts and hold significant semantic value; (3) Recognizing emojis in texts, as the emojis often convey abundant emotional information, crucial for rumor detection reliant on stance and sentiment recognition. 

\subsubsection{Auxiliary Task Limitations}

PLMs typically utilize various pretraining auxiliary tasks to enhance abilities in several aspects, including understanding sentence relations, recognizing entities, and managing grammatical rules. Mainstream auxiliary tasks encompass Next Sentence Prediction \cite{bert}, Sentence Order Prediction \cite{albert}, Replaced Token Detection \cite{ernie}, etc. These tasks carry out pretraining by learning semantic relations within documents, while rumor detection tasks focus more on interactive relations between documents (posts), particularly stance-related semantic relations. This is mainly because the content of a claim's comment is typically not independent but directional. Users generally express their opinions in response to content posted by other users.

\section{Method}

In this section, we introduce the datasets curated, and describe how we utilize PEP to train SoLM.

\subsection{Pretraining Corpora}

\textbf{TwitterCorpus} is a pure text Twitter corpus, which uses The Twitter Stream Grab publicly available on the Archive Team\footnote{\url{https://archive.org/details/twitterstream}} as its data source. It has extracted 2.8 billion English tweets from 2015 to 2022, totaling 269GB of uncompressed texts.

\textbf{UTwitter} contains trending claims from the past two years, collected from Twitter using a web crawler. It comprises about 200,000 unlabeled claims, each with a source post, multiple replies, and its propagation structure, totaling about 17 million tweets. Besides PLM pre-training, it can also be used for semi-supervised rumor detection.

\textbf{UWeibo} contains about 200,000 trending claims from Weibo over the past two years, with about 11 million posts.

TwitterCorpus, UTwitter, and UWeibo datasets are all available at \url{https://mega.nz/folder/wZwFGTzR#eAg4o-xJw3SBxfd2R3AmwQ}, \url{https://github.com/CcQunResearch/UTwitter}, and \url{https://github.com/CcQunResearch/UWeibo}.
The statistics and construction process are in Table~\ref{tab:sta} and Appendix~\ref{sec:databuild}. See Appendix~\ref{sec:pa} for dataset preprocessing and SoLM architecture.

\subsection{Post Engagement Prediction}

A claim conversation or propagation structure can be seen as a graph or, more specifically, a tree \cite{rvnn}.  This structure is characterized by a canonical node sorting within the tree, which proceeds either top-down or bottom-up \cite{bigcn}. Existing rumor detection methods based on propagation structures take advantage of the reply relation within the structures to learn the interaction of stance and sentiment between posts, thus identifying discriminative patterns to detect rumors. Thus, it is critical for PLMs to capture these interactive features between nodes in the trees. Yet, this is an aspect that current universal PLMs typically lack, as they tend to focus more on semantic connections within lengthy documents rather than modeling correlations between short ones, which is essential for social media application tasks such as rumor detection. Recognizing this issue, we propose the PEP strategy to assist PLMs in integrating the interaction information in propagation trees.

We found that nodes within a rumor propagation tree share certain connections, including: (1) All nodes are intrinsically linked to the root node, as all claim replies tend to revolve around the source post, discussing specific topics; (2) Nodes on the same branch form a conversation thread with closely related content, where deeper successor nodes are semantically dependent on the shallower prefix nodes; (3) Directly connected nodes exhibit a clear reply relation, with child nodes often stating explicit stances or sentiments towards parent nodes. Such clear semantic connections reflected via the graph structure are due to the claim propagation tree's canonical node sorting.

\begin{figure}[t]
  \centering
  \includegraphics[width=0.46\textwidth]{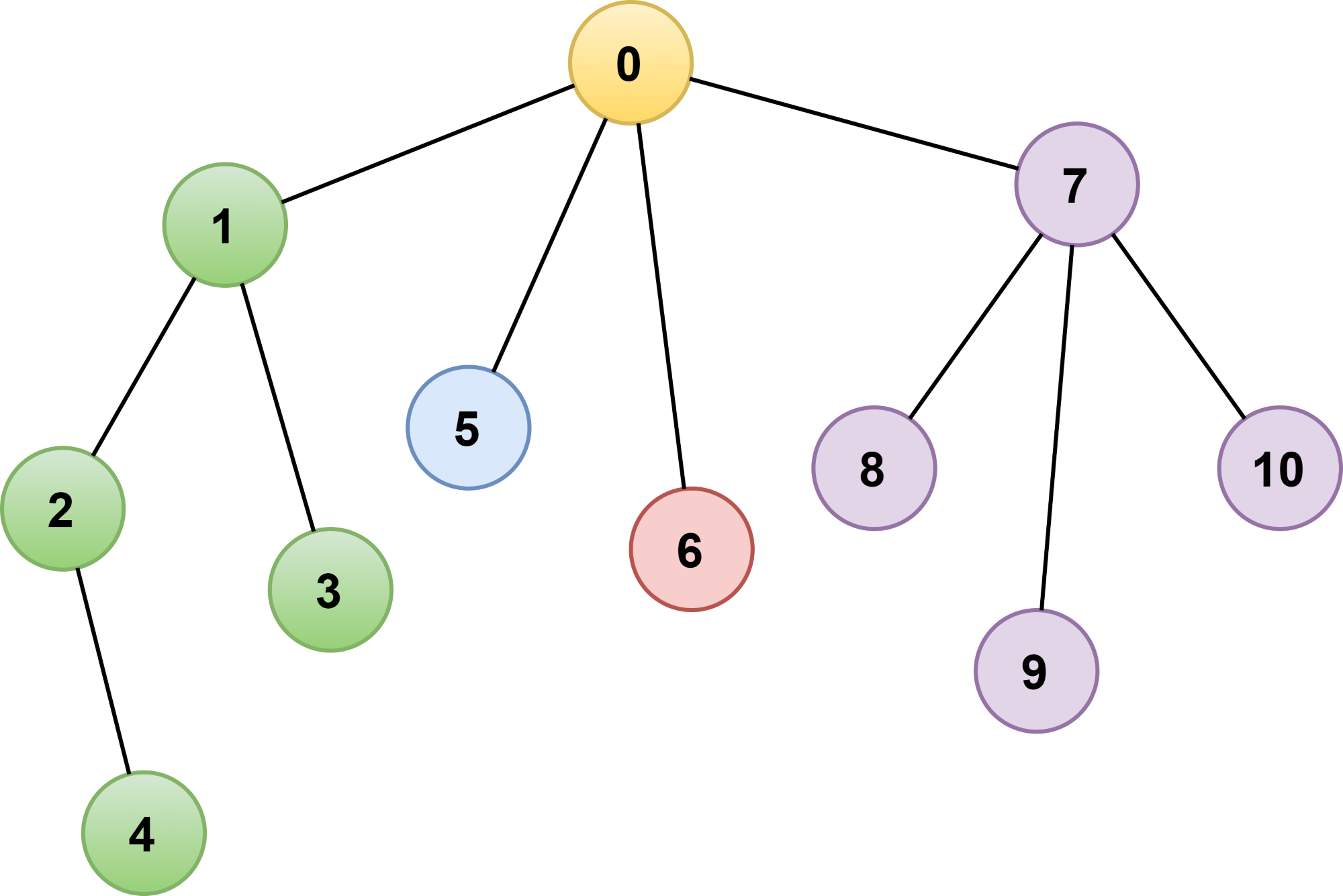}
  \caption{An example of rumor propagation tree. Different colors 
  correspond to different conversation threads.}
  \label{fig:rpt}
\end{figure}

PEP is a continue pretraining strategy that is conceptually straightforward in its formulation. It uses these node relations conveyed by the propagation structure as self-supervised information to assist PLM pretraining. Specifically, PEP includes Root Prediction (RoP), Branch Prediction (BrP), and Parent Prediction (PaP), which allow a PLM to predict the root, branch, and parent relations in propagation trees, respectively. For example, some RoP, BrP and PaP labels in Figure~\ref{fig:rpt} can be illustrated in Table~\ref{tab:label}. It is worth noting that although we use BERT as basic architecture of SoLM, PEP can also assist to pretrain all mainstream PLM architectures for tree-structured tasks such as rumor detection.

\begin{table}[t]
\centering
\begin{tabular}{cc|ccc}
\Xhline{1.0pt}
 \rowcolor{gray!20}
\textbf{Post1} & \textbf{Post2} & \textbf{RoP} & \textbf{BrP} & \textbf{PaP}\\
\hline
 0 & 1 & T & T & T\\
 0 & 2 & T & T & F\\
 1 & 2 & F & T & T\\
 1 & 4 & F & T & F\\
 1 & 7 & F & F & F\\
 4 & 7 & F & F & F\\
 .. & .. & .. & ..  & ..\\
\Xhline{1.0pt}
\end{tabular}
\caption{Root, branch and parent relation labels derived from the tree in Figure~\ref{fig:rpt}. T for True, F for False.}
\label{tab:label}
\end{table}

\noindent \textbf{Root Prediction.} RoP promotes learning interactions between a source post and its comment posts by predicting if two nodes are in a root relation (i.e., whether one node is the root node of another). Specifically, for a propagation tree $\mathcal{G}=(\mathcal{V},\mathcal{E})$, where $\mathcal{V}$ and $\mathcal{E}$ are the sets of nodes and edges. $\mathbf{H}_{\mathcal{G}}\in \mathbb{R}^{n\times d}$ is node feature matrix, where $n$ is node number, and $d$ is dimension of feature vectors. Each row vector in \(\mathbf{H}_{\mathcal{G}}\) represents a sentence embedding extracted from a PLM for a corresponding post. This could be, for instance, the embedding vector of the \texttt{[CLS]} token in BERT, or the embedding vector of the final token of a post in an autoregressive model. Then, the loss of RoP is:
\begin{equation}
\mathcal{L}_{\mathrm{RoP}}=-\frac{1}{|\mathbb{G}|}\sum _{\mathcal{G}\in \mathbb{G}}CE(\sigma (\mathbf{H}_{\mathcal{G}}\mathbf{H}_{\mathcal{G}}^{T}),\mathbf{Y}_{\mathrm{RoP}}),
\end{equation}
where $\mathbb{G}$ is the set of claim propagation trees corresponding to the claims in UTwitter or UWeibo, $CE(\cdot ,\cdot )$ is the cross-entropy loss, $\sigma (\cdot )$ is the sigmoid activation function, and $\mathbf{Y}_{\mathrm{RoP}}\in \mathbb{R}^{n\times n}$ is the self-supervised label matrix of root relations extracted from propagation trees.

\noindent \textbf{Branch Prediction.} BrP predicts whether two nodes come from the same conversation thread in a propagation tree (the nodes with the same color in Figure~\ref{fig:rpt}). Usually, posts in the same conversation thread discuss root post's content from the same perspective. BrP captures the interaction of nodes in the same branch by learning this kind semantic connection. Similarly, we obtain the loss $\mathcal{L}_{\mathrm{BrP}}$ through $\mathbf{H}_{\mathcal{G}}$ and label matrix $\mathbf{Y}_{\mathrm{BrP}}\in \mathbb{R}^{n\times n}$:
\begin{equation}
\mathcal{L}_{\mathrm{BrP}}=-\frac{1}{|\mathbb{G}|}\sum _{\mathcal{G}\in \mathbb{G}}CE(\sigma (\mathbf{H}_{\mathcal{G}}\mathbf{H}_{\mathcal{G}}^{T}),\mathbf{Y}_{\mathrm{BrP}}).
\end{equation}

\noindent \textbf{Parent Prediction.} PaP is similar to link prediction \cite{lp1,lp2}. In a propagation tree, parent-child nodes that are directly connected have clear stances and emotional relations semantically. PaP facilitates the model to learn node interaction that are directly connected in a propagation tree by predicting whether two nodes are directly connected (that is, whether one node is the parent of the other). We use $\mathbf{H}_{\mathcal{G}}$ and label matrix $\mathbf{Y}_{\mathrm{PaP}}\in \mathbb{R}^{n\times n}$ to derive loss $\mathcal{L}_{\mathrm{PaP}}$:
\begin{equation}
\mathcal{L}_{\mathrm{PaP}}=-\frac{1}{|\mathbb{G}|}\sum _{\mathcal{G}\in \mathbb{G}}CE(\sigma (\mathbf{H}_{\mathcal{G}}\mathbf{H}_{\mathcal{G}}^{T}),\mathbf{Y}_{\mathrm{PaP}}).
\end{equation}

The final loss function of PEP is as follows:
\begin{equation}
\mathcal{L}_{\mathrm{PEP}}=\alpha \cdot \mathcal{L}_{\mathrm{RoP}}+\beta \cdot \mathcal{L}_{\mathrm{BrP}}+\gamma \cdot \mathcal{L}_{\mathrm{PaP}}.
\end{equation}
We set $\alpha =\beta =\gamma =1$ in our experiments.

\subsection{Training Strategy}

We train on TwitterCorpus with Masked Language Modeling (MLM) to learn basic knowledge (first stage). Then, we train on UTwitter using MLM and PEP (second stage). The process is in Algorithm~\ref{alg:train}.

\begin{algorithm}[!h]
  \caption{Pretraining Strategy}
  \label{alg:train}
  \begin{algorithmic}[1]
    \Require initial parameter $\theta ^{(0)}$, training step $N$ of first stage, training step $M$ of second stage.
    \Ensure optimized model parameter $\theta ^{(N+M)}$.
    \State $//$ First stage: pretraining on TwitterCorpus. 
    \For {$n=1$ to $N$}
    \State Update $\theta ^{(n)}$: minimize $\mathcal{L}_{\mathrm{MLM}}$.
    \EndFor
    \State $//$ Second stage: pretraining on UTwitter. 
    \For {$m=1$ to M}
    \State Update $\theta ^{(N+m)}$: minimize $\mathcal{L}_{\mathrm{MLM}}+\mathcal{L}_{\mathrm{PEP}}$.
    \EndFor\\
    \Return $\theta ^{(N+M)}$.
  \end{algorithmic}
\end{algorithm}

\section{Experiments}

This section presents a evaluation on performance. 

\subsection{Experimental Settings}

We verify the enhancement effect on baseline methods (typical high-level rumor detection methods).

\subsubsection{Datasets}

We experimented on five benchmark datasets in Table~\ref{tab:sta}. 
PHEME is a class-imbalanced dataset, while others are class-balanced.
We reported macro F1 score on PHEME, and accuracy on the others.

\subsubsection{Baselines}

We replace the feature initialization modules of the following baseline methods with PLMs trained by PEP to verify its performance enhancement.

\textbf{PLAN} \cite{plan} is based on Transformer architecture. Its StA-PLAN version uses rumor propagation structure information.

\textbf{BiGCN} \cite{bigcn} utilizes two bidirectional GCN encoders and root node feature enhancement strategy to classify rumor.

\textbf{ClaHi-GAT} \cite{clahi} uses GAT on undirected graphs with sibling relations to model user interactions.

\textbf{GACL} \cite{gacl} uses contrastive learning and adversarial training to classify rumor.

\textbf{RAGCL} \cite{ragcl} is the current SOTA method on the benchmark datasets. It uses contrast learning with adaptive data augmentation.

These baseline methods all follow a unified framework: (1) Extract initial text features using PLMs or word2vec; (2) Further encode the extracted features using high-level models such as GNNs or Graph Transformer; (3) Train the model using training strategies like contrastive learning or adversarial training. In the experiments conducted in Table~\ref{tab:init} and Table~\ref{tab:res}, we only replaced the feature initialization module in (1) to explore its impact, keeping components in (2) and (3) unchanged.

We further access SoLM's capability to manage rumor detection tasks independently, without using any high-level model (\emph{SoLM Only} in Table~\ref{tab:res}). Specifically, we follow the GNN approach for graph classification tasks, performing pooling on feature vectors of all posts related to a claim. Each post's feature vector is extracted from the \texttt{[CLS]} token representation in SoLM. A linear classifier is then applied. More details of experimental settings is shown in Appendix~\ref{sec:expdetail}. The source code of PEP are available at \url{https://github.com/CcQunResearch/SoLM}.

\subsection{Results and Discussion}

In our experiments, we evaluated the impact of various universal PLMs such as RoBERTa, Baichuan2, LLaMA2, the social media specific PLM TwHIN-BERT, and SoLM on rumor detection methods. We used UWeibo to continue pretraining Chinese PLMs (RoBERTa-base-Chinese and Baichuan2) and UTwitter for others, in order to separately process benchmark datasets in Chinese and English. For earlier PLMs like BERT and RoBERTa, their vocabularies struggle to effectively handle special symbols such as emojis in social text (as mentioned in Section~\ref{sec:sps}). We utilize reserved tokens (\texttt{[unused]}) in BERT or infrequently used tokens in RoBERTa to represent these special symbols. The experimental results are presented in Table~\ref{tab:res} (see Appendix~\ref{sec:ppp2} for results on other baselines). 

\begin{table*}[t]
\centering
\begin{tabular}{clccccc}
 \Xhline{1.0pt}
 \rowcolor{gray!20}
 ~ & ~ & \multicolumn{5}{c}{\textbf{Dataset}}\\
 \cline{3-7}
 \rowcolor{gray!20}
 \multirow{-2}{*}{\textbf{Method}} & \multirow{-2}{*}{\textbf{Initialization}} & \textbf{Weibo} & \textbf{DRWeibo} & \textbf{Twitter15} & \textbf{Twitter16} & \textbf{PHEME}\\
 \hline
 \multirow{10}{*}{\textbf{PLAN}} & RoBERTa & 91.8 & 78.3 & 82.4 & 83.0 & 67.8 \\
 ~ & \qquad w/ PEP & 94.0(↑2.2) & 81.1(↑2.8) & 84.2(↑1.8) & 85.8(↑2.8) & 69.0(↑1.2) \\
 \cdashline{2-7}
 ~ & TwHIN-BERT & - & - & 82.8 & 84.3 & 69.5 \\
 ~ & \qquad w/ PEP & - & - & 84.7(↑1.9) & 86.0(↑1.7) & 71.7(↑2.2) \\
 \cdashline{2-7}
 ~ & Baichuan2 & 92.5 & 79.4 & - & - & - \\
 ~ & \qquad w/ PEP & 94.8(↑2.3) & 83.2(↑3.8) & - & - & - \\
 \cdashline{2-7}
 ~ & LLaMA2 & - & - & 83.4 & 84.0 & 70.2 \\
 ~ & \qquad w/ PEP & - & - & 85.0(↑1.6) & 86.6(↑2.6) & 71.9(↑1.7) \\
 \cdashline{2-7}
 ~ & SoLM(MLM) & - & - & 83.3 & 84.6 & 68.7 \\
 ~ & SoLM & - & - & 85.2(↑1.9) & 87.0(↑2.4) & 70.6(↑1.9) \\
 \hline
 \multirow{10}{*}{\textbf{BiGCN}} & RoBERTa & 93.8 & 87.2 & 83.8 & 87.3 & 70.5 \\
 ~ & \qquad w/ PEP & 95.0(↑1.2) & 89.7(↑2.5) & 85.6(↑1.8) & 88.5(↑1.2) & 72.0(↑1.5) \\
 \cdashline{2-7}
 ~ & TwHIN-BERT & - & - & 85.2 & 87.2 & 71.8 \\
 ~ & \qquad w/ PEP & - & - & 87.0(↑1.8) & 88.7(↑1.5) & 73.8(↑2.0) \\
 \cdashline{2-7}
 ~ & Baichuan2 & 94.0 & 88.7 & - & - & - \\
 ~ & \qquad w/ PEP & 95.6(↑1.6) & \textbf{90.4(↑1.7)} & - & - & - \\
 \cdashline{2-7}
 ~ & LLaMA2 & - & - & 85.1 & 87.0 & 72.0\\
 ~ & \qquad w/ PEP & - & - & 87.3(↑2.2) & 88.2(↑1.2) & 74.0(↑2.0) \\
 \cdashline{2-7}
 ~ & SoLM(MLM) & - & - & 85.0 & 87.3 & 70.8 \\
 ~ & SoLM & - & - & 86.6(↑1.6) & 89.2(↑1.9) & 73.2(↑2.4) \\
 \hline
 \multirow{10}{*}{\textbf{GACL}} & RoBERTa & 93.4 & 86.4 & 85.3 & 89.4 & 70.3 \\
 ~ & \qquad w/ PEP & 95.2(↑1.8) & 89.0(↑2.6) & 86.4(↑1.1) & 90.4(↑1.0) & 72.1(↑1.8) \\
 \cdashline{2-7}
 ~ & TwHIN-BERT & - & - & 85.8 & 88.8 & 71.4 \\
 ~ & \qquad w/ PEP & - & - & 86.9(↑1.1) & 90.0(↑1.2) & 73.5(↑2.1) \\
 \cdashline{2-7}
 ~ & Baichuan2 & 94.3 & 87.1 & - & - & - \\
 ~ & \qquad w/ PEP & \textbf{96.5(↑2.2)} & 90.8(↑3.7) & - & - & - \\
 \cdashline{2-7}
 ~ & LLaMA2 & - & - & 86.0 & 89.8 & 72.3 \\
 ~ & \qquad w/ PEP & - & - & 87.3(↑1.3) & \textbf{90.8(↑1.0)} & 73.9(↑1.6) \\
 \cdashline{2-7}
 ~ & SoLM(MLM) & - & - & 85.4 & 89.1 & 72.8 \\
 ~ & SoLM & - & - & \textbf{87.4(↑2.0)} & 90.6(↑1.5) & 74.5(↑1.7) \\
 \hline
 \textbf{SoLM Only} & - & - & - & 82.6 & 83.9 & 67.4 \\
 \hline
 \textbf{RAGCL(SOTA)} & RoBERTa & 96.2 & 89.4 & 86.7 & 90.5 & \textbf{76.8} \\
 \Xhline{1.0pt}
\end{tabular}
\caption{Experimental results on benchmark datasets. SoLM(MLM) refers to SoLM without second stage training.}
\label{tab:res}
\end{table*}

The experimental results indicate that the PEP strategy significantly enhances the performance of these PLMs in rumor detection tasks. Specifically, on the Weibo, DRWeibo, Twitter15, Twitter16, and PHEME datasets, performance improvements of 1.2-2.2\%, 1.7-3.7\%, 1.1-2.2\%, 1.0-1.9\%, and 1.5-2.4\% were achieved, respectively. The peak performance achieved on multiple datasets even surpassed the latest SOTA results \cite{ragcl}. Furthermore, the direct utilization of SoLM's features for rumor identification, without relying on high-level models, also yielded considerable results. These experimental findings highlight the critical importance of the text feature extraction module's ability to effectively learn the interactive features between posts for typical social media application tasks like rumor detection. Our PEP training strategy integrates the user engagement information embedded in the propagation structure of claims into the semantics of PLMs in a straightforward manner, resulting in performance gains without altering the high-level model.

In addition, according to the result in Table~\ref{tab:init}, the performance of PLMs is comparable to word2vec. A possible explanation is that rumor detection models are not particularly sensitive to feature extraction method, with the performance being primarily driven by high-level model. However, the performance improvements observed in Table~\ref{tab:res} underscore the equal importance of underlying feature extraction methods. Extracting better features can aid high-level models in learning more discriminative patterns, thereby achieving superior results.

\subsection{Ablation Study}
\label{sec:ab}

We investigated the impact of SoLM's two training stages and various training tasks on model performance using Twitter15 and Twitter16 with BiGCN. 
The outcomes, shown in Table~\ref{tab:ab}, reveal the positive effect of TwitterCorpus on the PLM model's performance, likely due to resolving prior corpora mismatch issues. The impact of the second stage training on model performance is also significant. RoP and PaP notably outperform BrP in impacting model performance (order of importance: PaP >= RoP > BrP), implying rumor detection's reliance on interactions between replies and source posts, as well as between directly replied posts.

\begin{table}[t]
\centering
\resizebox{0.48\textwidth}{!}{
\begin{tabular}{lcc}
 \Xhline{1.0pt}
 \rowcolor{gray!20}
 ~ & \textbf{Twitter15} & \textbf{Twitter16} \\
 \hline
 BiGCN w/ SoLM & 86.6 & 89.2 \\
 \hline
 w/o MLM pretraining & 85.8(↓0.8) & 88.1(↓1.1) \\
 \hdashline
 w/o PEP pretraining & 85.0(↓1.6) & 87.3(↓1.9) \\
 \quad w/o RoP & 85.8(↓0.8) & 88.3(↓0.9) \\
 \quad w/o BrP & 86.3(↓0.3) & 88.7(↓0.5) \\
 \quad w/o PaP & 85.6(↓1.0) & 88.1(↓1.1) \\
 \Xhline{1.0pt}
\end{tabular}
}
\caption{Ablation study on corpora and PEP strategy.}
\label{tab:ab}
\end{table}

Training a domain specific PLM for social media from scratch using a large-scale corpus like TwitterCorpus demands substantial computational resources (e.g., SoLM requires eight A800 80GB SXM GPUs for 14 days of training). In contrast, employing PEP strategy to continue pretraining an existing universal PLM such as RoBERTa requires only a single A800 80GB SXM GPU for one day, while still achieving a similarly notable improvement in performance (see Table~\ref{tab:res}). Under conditions of limited computational resources, PEP can serve as an alternative strategy for training social media specific PLMs. Furthermore, the assumptions underlying the design of PEP are universally applicable to social text and rely solely on easily accessible claim texts and propagation structure. 

\subsection{Few-shot Performance}

\begin{figure}[h]
  \centering
  \includegraphics[width=0.48\textwidth]{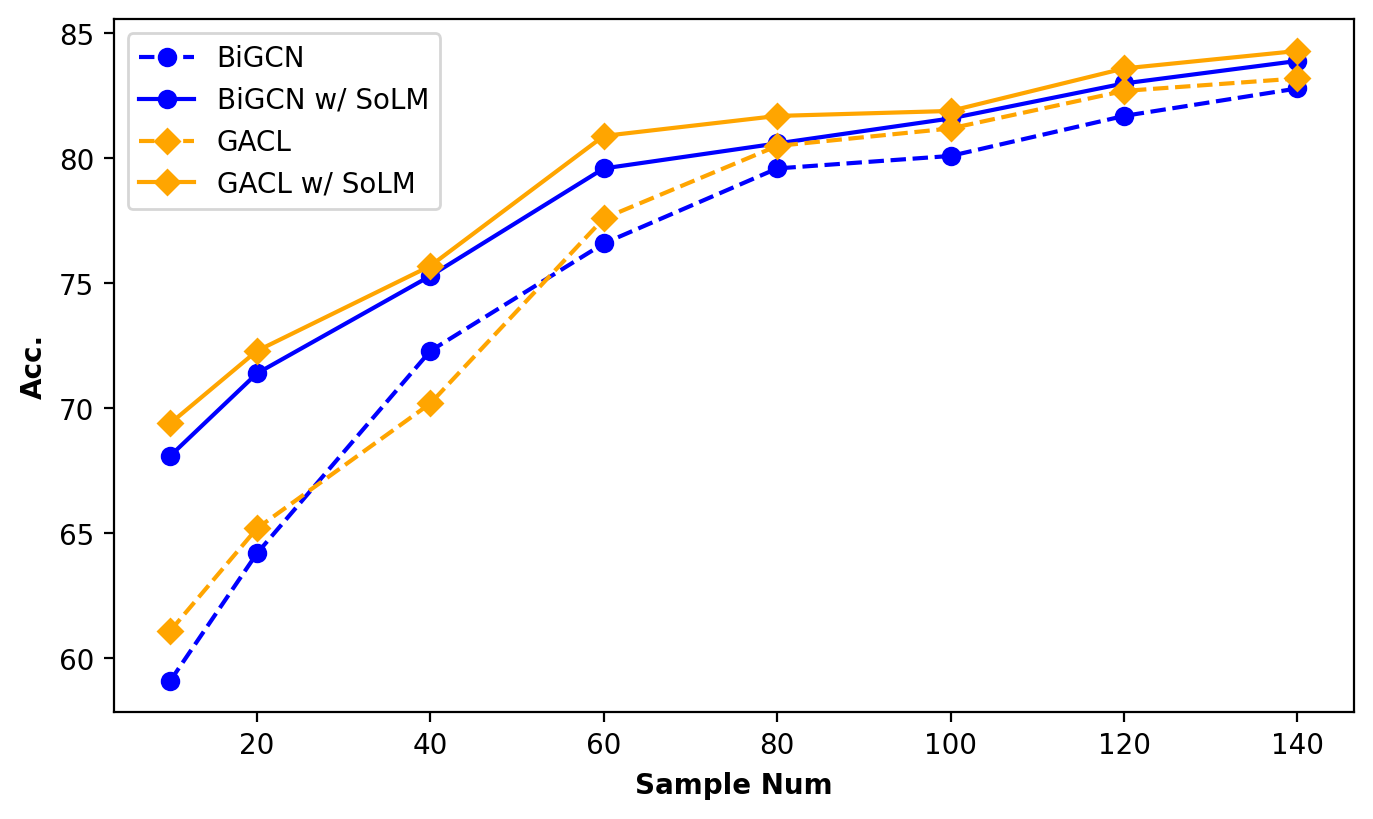}
  \caption{Results of few-shot experiments.}
  \label{fig:fs}
\end{figure}

As shown in Figure~\ref{fig:fs}, we use BiGCN and GACL to conduct few-shot learning experiments on Twitter15 to verify the enhancement effect of SoLM with only a few labeled samples. Because rumors are usually deleted after being detected, making it difficult to gain a large-scale labeled dataset, so the exploration of few-shot rumor detection is essential. We varied the number of labeled samples $k$ between 10 and 140. The results highlight that SoLM significantly enhances baseline model performance with fewer labeled samples. As the number of samples escalates, this enhancement effect tapers off. This superior few-shot performance indicates that SoLM has played a significant role in mitigating the overfitting issue in rumor detection models. See Appendix~\ref{sec:fs} for few-shot results on Twitter16.

\section{Conclusion}

In conclusion, this study identifies significant limitations in using universal PLMs for rumor detection and introduces a novel continue pretraining strategy, PEP, to address these issues. By pretraining on large-scale Twitter corpora and incorporating the PEP strategy focused on post interactions, our SoLM overcomes key deficiencies of traditional PLMs for this domain. Extensive experiments demonstrate that SoLM significantly enhances the performance of existing rumor detection methods, especially in few-shot scenarios.

\section*{Ethical Statement}

We employed web crawling tools to gather data from publicly available content posted by users on the Weibo and Twitter platforms. This content is accessible to any user of these platforms. To protect privacy, we will process the final dataset by removing any personally identifiable information, ensuring that no individual can be identified. Our exclusive aim in collecting and analyzing this data is for academic research, specifically to enhance the quality of information on social media and curb the spread of misinformation. By leveraging semi-supervised learning methods, we can improve model performance even with limited labeled data, contributing valuable insights to the field of rumor detection. Throughout our research, we are committed to upholding ethical standards, complying with legal requirements, and respecting our data, participants, and society at large.

\section*{Limitations}

In this study, we propose the PEP continue pretraining strategy and SoLM, which enhance the performance of PLMs in existing rumor detection models. However, these methods also have the following limitations:
(1) Although the underlying assumptions of the PEP strategy exhibit a certain degree of generality across different social media application tasks, we have only validated its performance on the rumor detection task so far. Further research is needed to explore its effectiveness in other tasks.
(2) The PEP strategy relies on data from specific platforms for pretraining. Although the PEP strategy exhibits a certain degree of cross-platform generalizability, the individual model trained does not possess this capability.
(3) Rumor detection faces challenges such as rapid updates, fast dissemination, and significant harm. The performance of the PEP strategy in tasks that require timely updates of information needs further validation.

\section*{Acknowledgments}

The authors would like to thank all the anonymous reviewers for their help and insightful comments.
This work is supported in part by the National Key R\&D Program of China (2018AAA0100302) and the National Natural Science Foundation of China (61876016).

\bibliography{custom}

\appendix

\section{Claim Propagation Trees}
\label{sec:rpt}

\begin{figure}[h]
  \centering
  \subfigure[Rumor]{\includegraphics[width=0.48\linewidth]{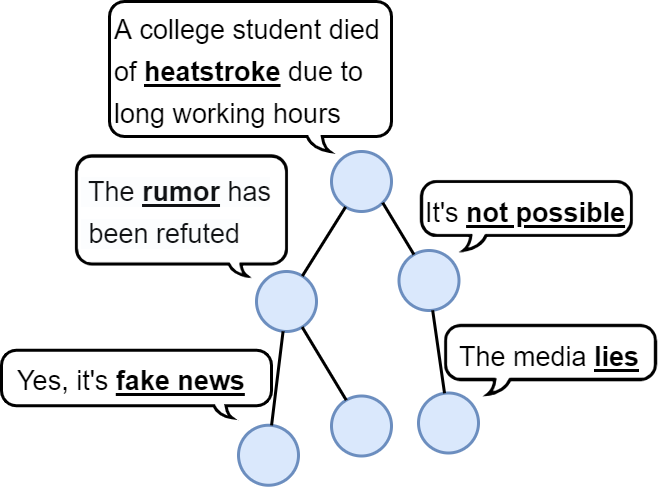}}
  \subfigure[Non-rumor]{\includegraphics[width=0.48\linewidth]{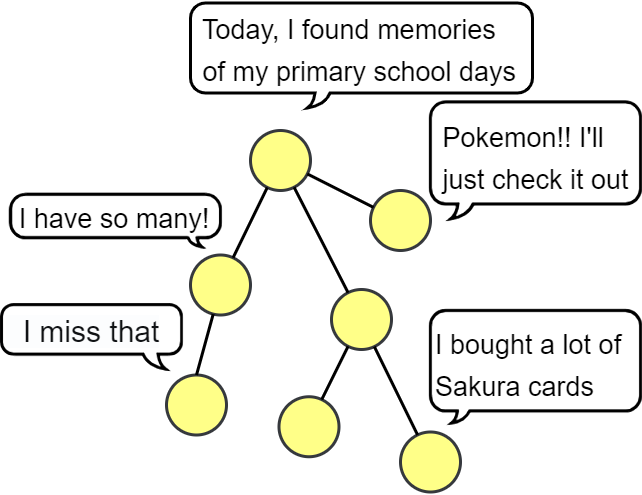}}
  \caption{Examples of claim propagation trees. Comments under rumor conversation typically express more heated stances and sentiments.}
  \label{fig:pt}
\end{figure}

Figure~\ref{fig:pt} shows two examples of claim propagation trees from the Twitter platform, where the replies of rumor and non-rumor claims exhibit distinct differences in stance and sentiment. These are key features for identifying rumors.

\section{Unlabeled Dataset Construction}
\label{sec:databuild}

For the UWeibo dataset, we employed web crawler techniques to randomly collect trending posts and their complete propagation structures from the homepage of popular Weibo posts\footnote{\url{https://weibo.com/hot/weibo/102803}}. To ensure the dataset's integrity and independence from platform recommendation algorithms, we utilized multiple newly created accounts to extract data. This approach aimed to mitigate potential biases that might arise from the platform's algorithms and to reflect the genuine domain distribution of social media content. The code for the web scraping program can be found at \url{https://github.com/CcQunResearch/WeiboPostAndCommentCrawl}.

For UTwitter dataset, we initially utilized multiple newly created accounts to randomly follow high-follower count influencers. Subsequently, we conducted random crawling of posts and their propagation structures from the Twitter homepage\footnote{\url{https://twitter.com/home}}. Due to the fact that UTwitter dataset is exclusively sourced from users with a substantial number of followers, the authenticity of the posts is more likely to be ensured. The code for the web scraping program is available at \url{https://github.com/CcQunResearch/TwitterPostAndCommnetCrawl}.

Due to the stringent regulations imposed by platforms on the dissemination of rumors, acquiring a sufficiently large-scale labeled dataset for rumor detection proves to be exceptionally challenging. Conversely, obtaining extensive amounts of unlabeled data is relatively simpler, especially with the availability of platform data APIs offered by certain mainstream social media platforms (e.g., Twitter API). Consequently, we suggest that future research should place greater emphasis on semi-supervised rumor detection methods.

Regarding the issue of potential data leakage, we believe that its impact is minimal in our study. This is because the Twitter15 and Twitter16 datasets include both source tweets and their corresponding comments, with each source tweet typically associated with dozens to hundreds of varying comments. In contrast, the TwitterCorpus dataset comprises only source tweets from the years 2015-2022 and does not include any comments. Therefore, the majority of the texts in Twitter15 and Twitter16 are not present in TwitterCorpus. As for UTwitter, it contains data from only the past one to two years, which does not overlap temporally with the Twitter15 and Twitter16 datasets (from the years 2015 and 2016). Consequently, we believe that data leakage is not a significant concern for the methodology employed in our paper.

\section{Experimental Details}
\label{sec:expdetail}

This section primarily details the experimental setup.

\subsection{Main Setting}

All models are implemented by PyTorch and the baseline methods are re-implemented. 
It should be noted that BiGCN and GACL utilize early stopping to observe the performance that models can achieve. However, due to oscillations in the early stages of model training, the observed model performance is unstable. In order to compare the performance of different models more fairly, we conduct experiments on multiple baseline methods with the same data, while all models are trained for 100 epochs until convergence. We consider the average results of the final 10 epochs out of these 100 as the stable outcome that the models can achieve.

\subsection{Preprocessing and Architecture}
\label{sec:pa}

For the texts in TwitterCorpus and UTwitter, we first standardize the different fonts present in the texts, then identify user mentions and web links as special tokens, \texttt{<@user>} and \texttt{<url>}. Next, we use the TweetTokenizer from the NLTK toolkit \cite{nltk} to tokenize the raw texts. Then, we use the \texttt{emoji} package\footnote{\url{https://pypi.org/project/emoji}} to translate the emojis in the texts into text string tokens.
Considering the vast scale of our corpora and the fact that tweets usually contain a lot of informal language, slang, internet jargon, and emojis, we set a larger vocabulary size of 52,000 for SoLM. Our SoLM adopts BERT\textsubscript{base} \cite{bert} as the model architecture. In conjunction with the previous statistics on text length, we set the maximum positional encoding of the model to 128. In total, there are seven special tokens in the vocabulary: \texttt{[UNK]}, \texttt{[SEP]}, \texttt{[PAD]}, \texttt{[CLS]}, \texttt{[MASK]}, \texttt{<@user>}, and \texttt{<url>}. 

\subsection{Optimization}

We use Huggingface Transformers \cite{transformers} to implement the basic architecture of SoLM. We set the maximum sequence length to 128 and use the AdamW optimizer \cite{adamw} to optimize the model. In the first stage of model training, we set the batch size to 8,000 and the peak learning rate to 0.0004, and use the first 4 epochs out of 40 epochs to warm up the learning rate. In the second stage, we set the batch size to 64, the peak learning rate to 0.00005, and use the first 2 epochs out of 20 epochs to warm up the learning rate. The entire training process was conducted over a span of 14 days on eight A800 80GB SXM GPUs. 

\section{Extended Experiments}

This section will present additional evaluation experiments.

\subsection{PEP Performance on Additional Baselines}
\label{sec:ppp2}

Similar to Table~\ref{tab:res}, we validated the performance of the PEP strategy and SoLM on ClaHi-GAT \cite{clahi} and RAGCL \cite{ragcl}, shown in Table~\ref{tab:res2}. Similarly, the PEP strategy enhances the performance of various PLMs in the rumor detection task, further confirming the generalizability of the PEP strategy. The performance improvement of the PEP strategy on ClaHi-GAT indicates that the language model trained with the PEP strategy is not only suitable for models of the Transformer architecture such as PLAN, as well as GCN architectures like BiGCN and GACL, but also effectively applicable to models of the GAT architecture. Our experiments in Appendix~\ref{sec:eas} further corroborate this. Additionally, as a recent SOTA method, RAGCL has already achieved excellent performance, but the language model trained with the PEP strategy provided a performance enhancement of 0.7-2.0\% across various datasets.

\begin{table*}[!h]
\centering
\begin{tabular}{clccccc}
 \Xhline{1.0pt}
 \rowcolor{gray!20}
 ~ & ~ & \multicolumn{5}{c}{\textbf{Dataset}}\\
 \cline{3-7}
 \rowcolor{gray!20}
 \multirow{-2}{*}{\textbf{Method}} & \multirow{-2}{*}{\textbf{Initialization}} & \textbf{Weibo} & \textbf{DRWeibo} & \textbf{Twitter15} & \textbf{Twitter16} & \textbf{PHEME}\\
 \hline
 \multirow{10}{*}{\textbf{ClaHi-GAT}} & RoBERTa & 93.4 & 86.4 & 85.0 & 88.5 & 70.3 \\
 ~ & \qquad w/ PEP & 94.8(↑1.4) & 88.1(↑1.7) & 86.6(↑1.6) & 89.9(↑1.4) & 72.5(↑2.2) \\
 \cdashline{2-7}
 ~ & TwHIN-BERT & - & - & 85.3 & 88.6 & 70.9 \\
 ~ & \qquad w/ PEP & - & - & 87.1(↑1.8) & 90.6(↑2.0) & 72.6(↑1.7) \\
 \cdashline{2-7}
 ~ & Baichuan2 & 94.0 & 86.9 & - & - & - \\
 ~ & \qquad w/ PEP & 95.1(↑1.1) & 89.4(↑2.5) & - & - & - \\
 \cdashline{2-7}
 ~ & LLaMA2 & - & - & 85.4 & 89.1 & 71.4 \\
 ~ & \qquad w/ PEP & - & - & 86.7(↑1.3) & 90.7(↑1.6) & 73.2(↑1.8) \\
 \cdashline{2-7}
 ~ & SoLM(MLM) & - & - & 85.6 & 88.9 & 72.1 \\
 ~ & SoLM & - & - & 87.3(↑1.7) & 90.9(↑2.0) & 74.2(↑2.1) \\
 \hline
 \multirow{10}{*}{\textbf{RAGCL}} & RoBERTa & 96.2 & 89.4 & 86.7 & 90.5 & 76.8 \\
 ~ & \qquad w/ PEP & 96.9(↑0.7) & 90.8(↑1.4) & 87.8(↑1.1) & 91.4(↑0.9) & 78.8(↑2.0) \\
 \cdashline{2-7}
 ~ & TwHIN-BERT & - & - & 86.4 & 90.3 & 77.0 \\
 ~ & \qquad w/ PEP & - & - & 87.4(↑1.0) & 91.6(↑1.3) & 78.6(↑1.6) \\
 \cdashline{2-7}
 ~ & Baichuan2 & 95.9 & 89.9 & - & - & - \\
 ~ & \qquad w/ PEP & 96.8(↑0.9) & 91.4(↑1.5) & - & - & - \\
 \cdashline{2-7}
 ~ & LLaMA2 & - & - & 86.6 & 90.3 & 77.1 \\
 ~ & \qquad w/ PEP & - & - & 87.7(↑1.1) & 91.3(↑1.0) & 78.9(↑1.8) \\
 \cdashline{2-7}
 ~ & SoLM(MLM) & - & - & 86.5 & 91.0 & 77.0 \\
 ~ & SoLM & - & - & 87.5(↑1.0) & 92.3(↑1.3) & 78.6(↑1.6) \\
 \hline
 \textbf{SoLM Only} & - & - & - & 86.7 & 90.5 & 76.8 \\
 \Xhline{1.0pt}
\end{tabular}
\caption{Experimental results of additional baseline models on benchmark datasets.}
\label{tab:res2}
\end{table*}

\subsection{Extended Ablation Study}
\label{sec:eas}

We conducted experiments on three commonly used GNN encoders, namely Graph Convolutional Network (GCN) \cite{gcn}, Graph Attention Network (GAT) \cite{gat}, and Graph Isomorphism Network (GIN) \cite{gin}, to explore the generality of SoLM in enhancing the performance of various high-level models. The experimental results are presented in Table~\ref{tab:gnn}, indicating that SoLM consistently improves the performance of these generic GNNs in rumor detection tasks. This observation underscores the versatility of SoLM across different models for rumor detection.

\begin{table}[!h]
\centering
\resizebox{0.48\textwidth}{!}{
\begin{tabular}{cccc}
 \Xhline{1.0pt}
 \rowcolor{gray!20}
 \textbf{Method} & \textbf{Initialization} & \textbf{Twitter15} & \textbf{Twitter16} \\
 \hline
 \multirow{2}{*}{GCN} & RoBERTa & 81.5 & 83.3 \\
  & SoLM & 83.5(↑2.0) & 84.7(↑1.4) \\
 \hline
 \multirow{2}{*}{GAT} & RoBERTa & 80.9 & 82.1 \\
  & SoLM & 83.0(↑2.1) & 83.8(↑1.7) \\
 \hline
 \multirow{2}{*}{GIN} & RoBERTa & 81.9 & 82.9 \\
  & SoLM & 84.2(↑2.3) & 84.4(↑1.5) \\
 \Xhline{1.0pt}
\end{tabular}
}
\caption{Performance enhancement on general GNNs.}
\label{tab:gnn}
\end{table}

\subsection{Extended Few-shot Experiments}
\label{sec:fs}

\begin{figure}[!h]
  \centering
  \includegraphics[width=0.48\textwidth]{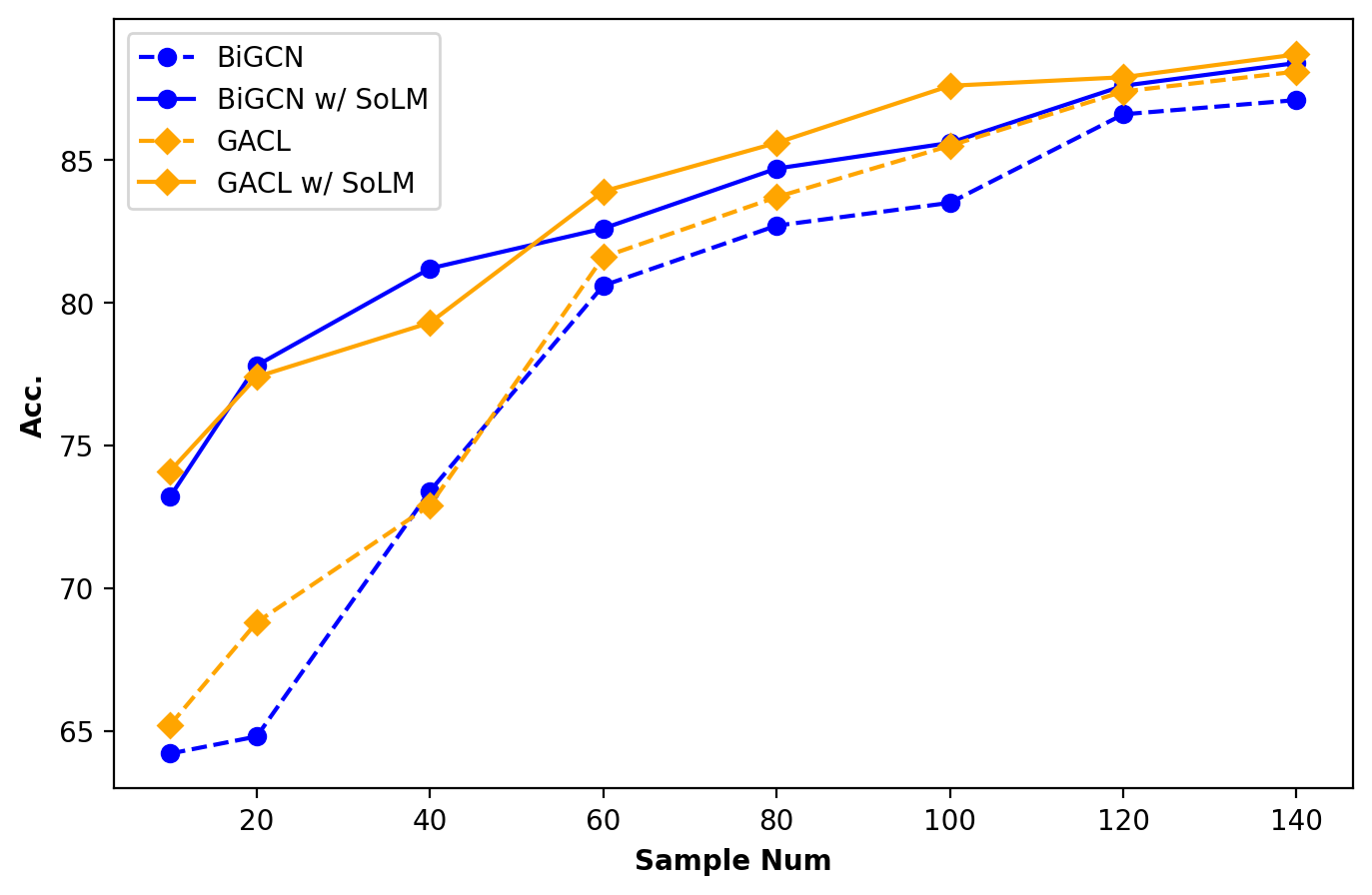}
  \caption{Results of few-shot experiments on Twitter16 dataset.}
  \label{fig:fs2}
\end{figure}

We present few-shot experiments on the Twitter16 dataset in Figure~\ref{fig:fs2}. Similar to the results in Figure~\ref{fig:fs}, pre-training SoLM on large-scale data improves the performance of existing rumor detection models when confronted with a small number of labeled samples.

\section{Discussions}

In this section, we will address some concerns that readers may have regarding the PEP strategy and the SoLM language model.

\subsection{Consumption of Computing Resources}

As previously discussed in Section~\ref{sec:ab}, although pre-training on the TwitterCorpus requires substantial computational resources (8 A100 GPUs for 14 days), fortunately, fine-tuning an existing open-source language model using the PEP strategy requires only a single A100 GPU for one day. This is a manageable requirement for most developers and still results in significant performance improvements (as shown in Table~\ref{tab:res} and Table~\ref{tab:res2}). We also highly recommend utilizing the method of fine-tuning open-source models. In fact, the comparative experiments on the performance of SoLM(MLM) and SoLM in Tables~\ref{tab:res} and~\ref{tab:res2} also indicate that the PEP continue training process in the second stage, which integrates propagation structure information, is more important than training a language model from scratch.

\subsection{Platform Generalizability}

It is important to note that the cross-platform generalizability we focus on refers to the ability of the PEP strategy to utilize data from different platforms for training and to be effective across those respective platforms, rather than training a single model that performs effectively on any platform's application task. This distinction is made because, in real-world applications, the latter is relatively meaningless.

Different platforms indeed exhibit noticeable differences, likely due to the varying user bases they cater to (across different age groups, ethnicities, languages, etc.), as well as the influence of platform-specific recommendation algorithms \cite{gacl,ragcl}. However, these platforms (such as mainstream platforms like Twitter, Weibo, Reddit, YouTube and TikTok) typically organize post responses in a tree structure \cite{rvnn,bigcn}. Our PEP strategy does not make platform-specific assumptions but rather utilizes the topological relations among nodes within this tree structure. The PEP strategy leverages unlabeled data and self-supervised learning to capture language patterns, thus possessing a certain degree of adaptability across different platforms. 

Our PEP strategy merely leverages the primary topological relations within the tree structure (Root, Branch, Parent Relation) for self-supervised LM continue pretraining. Our training approach only adds a linear layer to the model to predict the node topological relations and does not employ contrastive learning \cite{gacl}, attention mechanisms \cite{clahi}, or other strong prior assumption methods \cite{sog,ebgcn,ragcl}. This is primarily to minimize unnecessary inductive biases and to make as few prior assumptions about the data as possible. We aim for the model to learn the general relations between replies rather than specific behaviors. 

Experimental results have demonstrated that PEP consistently shows effectiveness on both Twitter and Weibo. Therefore, it can be concluded that the PEP strategy exhibits a certain degree of generalizability across different platforms. As for its performance beyond Twitter and Weibo, further research can be conducted in the future.

\end{document}